\documentclass[letterpaper, 10 pt, conference]{ieeeconf}  

\IEEEoverridecommandlockouts                              

\usepackage{amsmath} 
\usepackage{bm}
\usepackage{comment}
\usepackage{amssymb}
\usepackage{subcaption}
\usepackage{graphicx}
\usepackage{verbatim}
\usepackage{multirow, makecell}
\usepackage{hhline}
\usepackage{balance}
\usepackage{array, booktabs}
\usepackage[font={small}]{caption}
\usepackage[hidelinks]{hyperref}
\usepackage{soul}

\usepackage{cite}
\title{\LARGE \bf
Interaction-aware Decision Making with Adaptive Strategies under Merging Scenarios
}

\author{Yeping Hu$^{1}$, Alireza Nakhaei$^{2}$, Masayoshi Tomizuka$^{1}$, and Kikuo Fujimura$^{2}$
\thanks{$^{1}$Y. Hu and M. Tomizuka are with the Department of Mechanical Engineering, University of California, Berkeley, CA 94720 USA
        {\tt\small [yeping\_hu, tomizuka@berkeley.edu]}}%
\thanks{$^{2}$A. Nakhaei and K. Fujimura are with the Honda Research Institute, Mountain View, CA 94043 USA
        {\tt\small [anakhaei, kfujimura@hri.com]}}%
}

\makeatletter
\def\endthebibliography{%
	\def\@noitemerr{\@latex@warning{Empty `thebibliography' environment}}%
	\endlist
}
\makeatother
\begin{document}

\maketitle
\thispagestyle{empty}
\pagestyle{empty}

\begin{abstract}
In order to drive safely and efficiently under merging scenarios, autonomous vehicles should be aware of their surroundings and make decisions by interacting with other road participants. Moreover, different strategies should be made when the autonomous vehicle is interacting with drivers having different level of cooperativeness. Whether the vehicle is on the merge-lane or main-lane will also influence the driving maneuvers since drivers will behave differently when they have the right-of-way than otherwise. Many traditional methods have been proposed to solve decision making problems under merging scenarios. However, these works either are incapable of modeling complicated interactions or require implementing hand-designed rules which cannot properly handle the uncertainties in real-world scenarios.
In this paper, we proposed an interaction-aware decision making with adaptive strategies (IDAS) approach that can let the autonomous vehicle negotiate the road with other drivers by leveraging their cooperativeness under merging scenarios. A single policy is learned under the multi-agent reinforcement learning (MARL) setting via the curriculum learning strategy, which enables the agent to automatically infer other drivers' various behaviors and make decisions strategically. A masking mechanism is also proposed to prevent the agent from exploring states that violate common sense of human judgment and increase the learning efficiency. An exemplar merging scenario was used to implement and examine the proposed method.

\end{abstract}

\section{Introduction}
Recent advanced driving assistant systems (ADAS) have had functionalities such as adaptive cruising control (ACC) and automatic parking, however, autonomous vehicles are still unable to exhibit human-like driving behaviors which require interactive decision-making. In fact, to achieve general intelligence, agents must learn how to interact with others in a shared environment. 
\begin{figure}[htbp]
	\centering
	\includegraphics[scale=0.2]{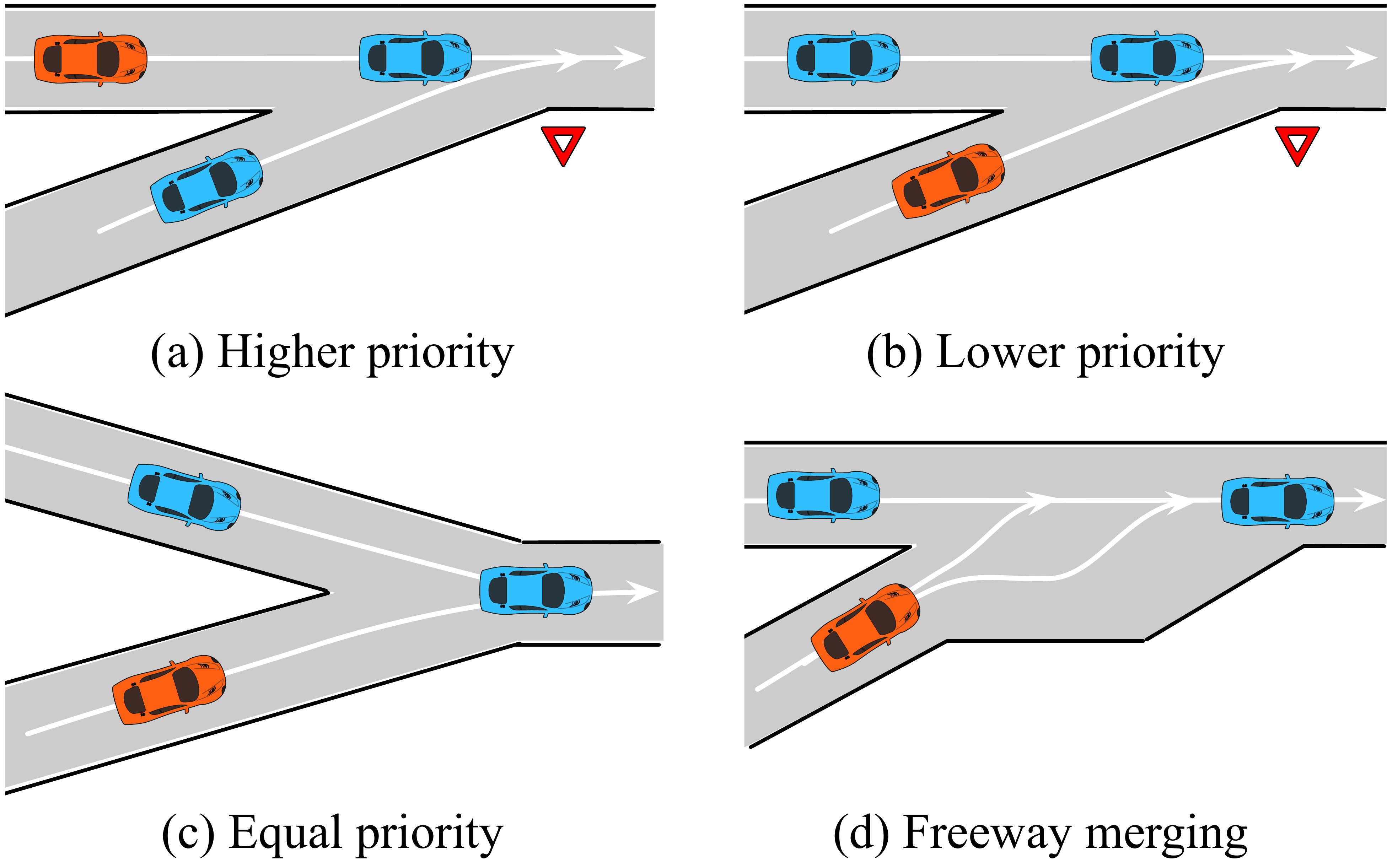}
	\caption{Illustrations of various merging scenarios. Red car represents the autonomous vehicle. In (a) and (b), cars with higher priorities have the right of way. However, some level of negotiations is necessary if the traffic gets dense. In (c), the car which is closer to the merge point has the right of way. However, negotiation is required since sometime it is difficult to define who is closer and each driver may have different opinions. Different from other scenarios, the red car in (d) has unknown merge point during merging. }
	\label{fig:merge}
\end{figure}

\subsection{Problem Description}
For this work, our goal is to solve decision making problems under merging scenarios. In general, merging scenarios can be divided into two categories shown in Fig.~\ref{fig:merge}, which contain the ones with unknown merge point such as highway entrance and the ones with known merge point that are mostly seen in urban areas. In our problem setting, only the latter category is considered, where all vehicles' motions will be on preplanned paths.

Nowadays, the industry has been taking conservative approaches to deal with merging scenarios for autonomous vehicles. The decision making is mainly based on the road priorities (which is defined in the traffic regulation) and rule-based prediction such as constant velocity. In these methods, the decision making algorithms do not rely on cooperativeness of other drivers on roads which causes autonomous vehicles make suboptimal decisions. If merging roads have similar priority as in Fig.1 (c), it is unreasonable to let the self-driving cars yield at all time and thus it needs to learn how to negotiate with other drivers. Moreover, without considering the cooperativeness of other drivers, autonomous vehicles may never merge on to dense traffic on main roads with higher priority. In this paper, we would like to leverage the cooperativeness of human drivers to negotiate the road with them. Specifically, we hope to generate maneuvers for self-driving cars which are safe but less conservative. 


\subsection{Challenges}
Although we focus on the known merge-point scenarios as in Fig. 1(a, b, c) in this work, it is still non-trivial for the autonomous vehicle to interact smoothly with human-driven cars. One of the challenges comes directly from the real-world scenarios: \textit{drivers on the road can behave differently from the cooperativeness point of view}, which makes the decision making difficult for self-driving cars. Other challenges come from the implementation aspect. For example, \textit{the driver model for other cars are usually unknown}, which is problematic since we need other cars' driver model to fine-tune and evaluate our algorithm against it so that we can develop a robust decision-making algorithm. This is in fact a cause and effect dilemma. Moreover, \textit{the current decision-making approaches cannot achieve a desired generalization}. In order for a decision maker to be used on self-driving cars, it needs to deal with various merging scenarios instead of one. The vehicle should be able to drive in a constantly changing environment, where the road priority, traffic density, and driving behaviors of surrounding road entities will vary.

\subsection{Related Works}
1) \textit{Decision Making:} Several rule-based and model-based methods \cite{model_3, model_4, model_5, model_1, model_2} have been utilized to solve the decision making problems under merging scenarios. In DARPA Urban Challenge, the winner utilized  a rule-based behavior generation mechanism \cite{model_1} for predefined traffic rules. \cite{model_2} modeled the decision process through a mixed logical dynamical system which is solved via model predictive control. Although these methods have reliable performances, they are only designed for specific domains and fail to model interactions between multiple road entities, which is a key component for navigating dynamic traffic environments efficiently and safely in the real-world.


In \cite{IRL}, the authors used Inverse Reinforcement Learning (IRL) to model how agents react to the robot's action. Despite interactions are considered, this approach assumes a full control over other vehicles with known models, which may lead the autonomous vehicle to aggressive and potentially dangerous behavior due to overconfidence in the behavior model \cite{review_paper}. Other work such as \cite{1D} used POMDP to solve decision making problem for autonomous driving, however, the interaction model they designed consists only a simple constant braking action that will be activated on emergency.

2) \textit{Multi-agent Reinforcement Learning (MARL):} A variety of MARL algorithms have been proposed to accomplish tasks without pre-programmed agent behaviors. Unlike single agent RL, where each agent observes merely the effect of its own actions, MARL enables agents to interact and learn simultaneously in the same environment. By utilizing MARL methods, agents can automatically learn how to interact with each other without any improper hand-design rules. 

A multi-agent deep deterministic policy gradient (MADDPG) method is proposed in \cite{MADDPG}, which learns a separate centralized critic for each agent, allowing each agent to have different reward functions. However, it does not consider the issue of multi-agent credit assignment which concerns with how the success of an overall system can be attributed to the various contributions of a system's components \cite{credit_assignment}. Such problem is addressed by Foerster et al. in \cite{COMA} and they proposed a counterfactual multi-agent (COMA) policy gradients that utilizes an actor and a centralized critic to solve cooperative multi-agent problems. A recent work by Yang et al. \cite{CM3} proposed a cooperative multi-goal multi-stage multi-agent reinforcement learning (CM3) approach, which learns an actor and a double critic shared by all agents. It not only takes the credit assignment issue into account as in \cite{COMA}, but also adopts the framework of centralized training with decentralized execution that enables the policy to use extra information to ease training. 


Although the aforementioned actor-critic methods have good performances in some tested simulation environments under the MARL setting, they consider only a single level of cooperativeness during interaction . Such problem settings cannot be utilized in real driving situations since various degrees of interaction exist between vehicles due to their different cooperativeness level. Also, while driving, the vehicle should not only focus on interacting with other road entities, but also pay attention to the overall traffic flow to not impede others. Therefore, we proposed a modified actor-critic method that takes both the microscopic and macroscopic-level interactions into account.

Moreover, the MARL learning process will be inefficient and slow if each agent explores every possible states including those that are implausible. In \cite{Mobileye2}, a responsibility-sensitive safety (RSS) model for multi-agent safety is developed, which is a formalism of the common sense of human judgment. We integrated such idea into our algorithm by introducing a masking mechanism similar to \cite{Safe_RL}, which prevents autonomous vehicles from learning those common sense rules. 

\subsection{Contributions}

In this paper, an interaction-aware decision making with adaptive strategies (IDAS) approach is proposed for autonomous vehicles to leverage the cooperativeness of other drivers in merging scenarios, which is built upon the multi-agent reinforcement learning (MARL). The contributions of this work are as follows: 

1) Utilizing the multi-agent setup to address the lack of driver model for other agents.

2) Introducing the notion of driver type to address the randomness of other drivers' behaviors.

3) Learning a single policy to obtain a generalizable decision maker that can apply adaptive strategies under various merging situations.

4) Incorporating both microscopic and macroscopic perspectives to encourage interactions and cooperativeness.

5) Leveraging curriculum learning and masking mechanism to improve learning efficiency.

With the proposed MARL method, each agent will have opportunities to interact with various types of surrounding agents during training and will be encouraged to learn how to select proper strategies when dealing with changing environments. In general, we want the self-driving cars implicitly reason about the cooperativeness of other drivers during interaction and exploit their willingness of cooperation to handle merging scenarios in a strategic way. 


\section{Preliminaries}
\subsection{Concept Clarification}
We introduced two variables throughout the paper: \textit{road priority} and \textit{driver type}. They will be used as input to the decision maker and the same policy is expected to behave differently by varying these two values.

1) \textit{Road priority}: It is a metric defined by traffic law in each city and the value is discrete. Decision maker does not need to do any assumption about it since it is available in the maps. 

2) \textit{Driver type}: It is a metric to define the cooperativeness of the decision maker and the value is continuous. Each agent in the environment will be randomly assigned with a fixed driver type and this information is not shared across agents. 


\subsection{Markov Games}
For this problem, we consider a multi-agent Markov game with $N$ agents labeled by $n \in [1,N]$.
The Markov game is defined by a set of states $\mathcal{S}$ describing the possible configurations of all agents, a set of partial observations $\mathcal{O}^n$ and a set of actions $\mathcal{A}^n$ for each agent. The road priority and driver-type information $\{ b_{prio}^n,b_{type}^n \} \in \mathcal{B}^n $ are two predetermined and fixed parameters that could influence the agent's behavior. Each agent $n$ chooses actions according to a stochastic policy $\pi^n: \mathcal{O}^n \times \mathcal{B}^n \times \mathcal{A}^n \rightarrow [0,1]$, and the joint action  of $N$ agents move the environment to the next state according to the transition function $\mathcal{T}: \mathcal{S} \times \mathcal{A}^1 \times ... \times \mathcal{A}^N \rightarrow \mathcal{S}$. Each agent receives a reward $r^n: \mathcal{S} \times \mathcal{B}^n \times \mathcal{A}^n \rightarrow \mathbb{R}$, which is a function of the state, agent's behavior, and agent's action, to maximize its own total expected return $R^n = \sum_{t=0}^{T} \gamma^t r^n_t$, where $\gamma \in [0,1)$ is a discount factor and $T$ is the time horizon.

\subsection{Actor-Critic Methods}

Actor-critic methods \cite{actor_critic} are widely used for variety of RL tasks in both single-agent and multi-agent environments. The actor is a parameterized policy that defines how actions are selected; the critic is an estimated state-value function that criticizes the actions made by the actor. The parameters of the actor are then updated with respect to the critic's evaluation. Different from the REINFORCE \cite{REINFORCE} algorithm that uses a stationary baseline value to reduce high variance gradient estimate, actor-critic methods use past experiences to update the baseline known as the critic. The learned critic components can shift the value of the critique enabling the gradient to point in the right direction. 

\subsubsection{Single agent}
In a single-agent setting, policy $\pi$, parametrized by $\theta$, maximizes the objective $J(\theta) = \mathbb{E}_\pi[R]$ by taking steps in the direction of $\nabla_\theta J(\theta)$, where the expectation $\mathbb{E}_\pi$ is with respect to the state-action distribution induced by $\pi$. The gradient of the policy can be written as:
\begin{equation}
	\nabla_\theta J(\theta) = \mathbb{E}_\pi\bigg[\sum_{t}\nabla_\theta \textrm{log} \pi(a_t|s_t)(Q^\pi(s_t,a_t) - b(s_t))\bigg],
\end{equation}
where $Q^\pi(s_t,a_t) = \mathbb{E}_\pi[\sum_{t'=t}^{T}\gamma^{t'} r(s_{t'},a_{t'})|s_t, a_t]$ is the action-value function for policy $\pi$, $b(s_t)$ is the introduced baseline, and their difference is known as the advantage function $A^\pi(s_t,a_t)$. By choosing the value function $V^\pi(s_t)$ as the baseline and using the temporal difference (TD) error as an unbiased estimate of the advantage function, we can rewrite the advantage function as $A^\pi(s_t,a_t) \approx r(s_t,a_t)+\gamma V^\pi(s_{t+1})-V^\pi(s_t) $.

\subsubsection{Multi agent}
For the multi-agent environment, a paradigm of \textit{Centralized Critic, Decentralized Actor} is proposed \cite{MADDPG}\cite{COMA}, which is an extension of actor-critic methods. The critic is augmented with full state-action information about the policies of other agents, while the actor only has access to local information. Moreover, to address the \textit{credit assignment} problem in multi-agent settings, which could enable each agent to deduce it's own contribution to the team's success, a counterfactual baseline
\begin{equation}
b(s,\bm{a}^{-n}) = 	\sum_{a'^{n}}\pi^n(a'^n|o^n)Q(s,(\bm{a}^{-n},a'^n))
\end{equation}
is suggested in \cite{COMA}, where it marginalizes out the actions $a$ of an agent $n$ and allows the centralized critic to reason about the counterfactuals in which only agent $n$'s actions change.



\section{Approach}
\subsection{Interaction-aware Decision Making}
Our goal is to solve the interaction-aware decision-making problem for autonomous vehicles using MARL strategy. In particular, we proposed to learn a single actor that can generate various driving behaviors and adaptive interaction strategies, as well as a pair of decentralized and centralized critics shared by all agents. We first show two learning objectives, corresponding to the need for agents to drive through merging scenarios while strictly following the rules, and interact with other agents for a more efficient merging while maintaining a good traffic flow. Then the combined objective is illustrated. 

1) Since each agent is assigned with different individual rewards in order to learn distinct behaviors, it is difficult to extract various learning signals from a joint reward and thus a decentralized critic for every agent with shared parameters is needed. The \textbf{decentralized critic} aimed to provide a policy gradient for agent to learn how to drive under merging scenarios by strictly following the rules while having different behaviors. The agent is not expected to react to other agents and it will only learn how to execute rational actions to finish its own task. By naming this first objective as $J_1$, the policy gradient is given by:
\begin{equation}
\begin{split}
	\nabla_\theta J_1(\theta) &\approx \mathbb{E}_\pi\bigg[\sum_{n=1}^{N}\sum_{t}\nabla_\theta \textrm{log}\pi(a_t^n|o_t^n,b^n)\big(r(o_t^n,a_t^n \\ 
	& \quad ,b^n) + \gamma V^\pi_{\phi_1}(o_{t+1}^n,b^n)-V^\pi_{\phi_1}(o_t^n,b^n)\big)\bigg],
\end{split}
\end{equation}
where $V^\pi_{\phi_1}(o_t^n,b^n)$ is the decentralized critic parametrized by $\phi_1$ and is updated by minimizing the loss:
\begin{equation}
\begin{split}
	\mathcal{L}(\phi_1) &= \frac{1}{2}\sum_{i} \bigg\Vert r(s_{i,t},a_{i,t}^n,b^n_i)+\gamma V^\pi_{\hat{\phi_1}}(o^n_{i,t+1},b^n_i) \\
	&   \quad- V^\pi_{\phi_1}(o^n_{i,t},b^n_i) \bigg\Vert ^2,
\end{split}
\end{equation} 
where $i$ is the number of sampled batches and $V^\pi_{\hat{\phi_1}}$ is the target network with parameters $\hat{\phi_1}$ that slowly update towards $\phi_1$. The target network is used to stabilize the training process.

2) Although strictly following the rules will result in no accident for merging scenes, it is not the optimal strategy if we consider macroscopic-level factors such as traffic flow rate. The \textbf{centralized critic} encourages each agent to interact with each other in order to have a joint success and maintain a smooth traffic. We name the second objective as $J_2$ and derive its policy gradient:

\begin{equation}
\begin{split}
	\nabla_\theta J_2(\theta) &= \mathbb{E}_\pi \bigg[\sum_{n=1}^{N}\nabla_\theta \textrm{log}\pi(a^n|o^n,b^n)(Q^\pi_{\phi_2}(s,\bm{a},\bm{b}) \\
	& -\sum_{a^{\prime n}} \pi^n (a^{\prime n}|o^n,b^n)Q^\pi_{\phi_2}(s,(\bm{a}^{-n},a^{\prime n}),\bm{b}))\bigg],
\end{split}
\end{equation}
where we utilize the counterfactual baseline discussed in the previous section and define the centralized critic as $Q^\pi_{\phi_2}(s,\bm{a},\bm{b}) = \mathbb{E}_\pi[\sum_{t'=t}^{T}\sum_{n=1}^{N}\gamma^{t'} r(s_{t'},a_{t'}^n,b^n)|s_t, a_t^n,b^n]$ by considering a joint reward of all agents. Parameterized by $\phi_2$, the centralized critic is updated by minimizing the loss:
\begin{equation}
\begin{split}
	\mathcal{L}(\phi_2)  &= \frac{1}{2}\sum_{i} \bigg\Vert \sum_{n=1}^{N}r(s_{i,t},a_{i,t}^n,b^n_i) \\
	& +\gamma Q^{\hat{\pi}}_{\hat{\phi_2}}(s_{i,t+1},\bm{\hat{a}}_{i,t+1},\bm{b}) - Q^\pi_{\phi_2}(s_{i,t},\bm{a}_{i,t},\bm{b}) \bigg\Vert ^2,
\end{split}
\end{equation}
where $\hat{\pi}$ denotes the target policy network and $\hat{\phi_2}$ represents parameters of the target centralized critic network. 

The overall policy gradient can thus be defined as:
\begin{equation}
	\nabla_\theta J(\theta) = \alpha\nabla_\theta J_1(\theta) + (1-\alpha)\nabla_\theta J_2(\theta),
\end{equation}
where $\alpha \in [0,1]$ is the weighting factor for the two objectives. By considering two separate objectives, we can easily use curriculum learning strategy during the training process, which will be introduced in the next section.
\begin{figure}[htbp]
	\centering
	\begin{subfigure}{0.5\textwidth}
		\centering
		\includegraphics[scale=0.65]{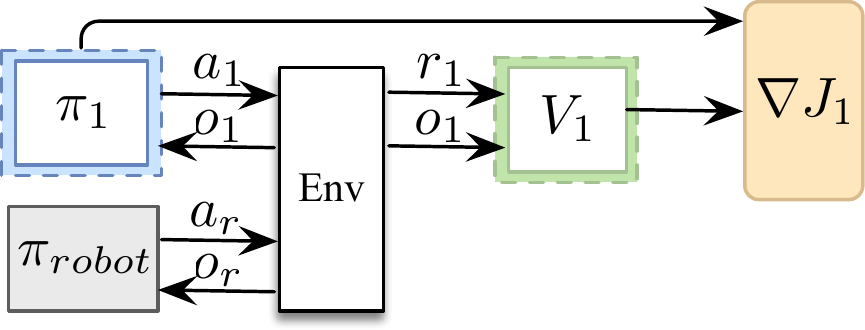}
		\caption{Stage1: Robot World}
	\end{subfigure}
	~ 
	\begin{subfigure}{0.5\textwidth}
		\centering
		\includegraphics[scale=0.65]{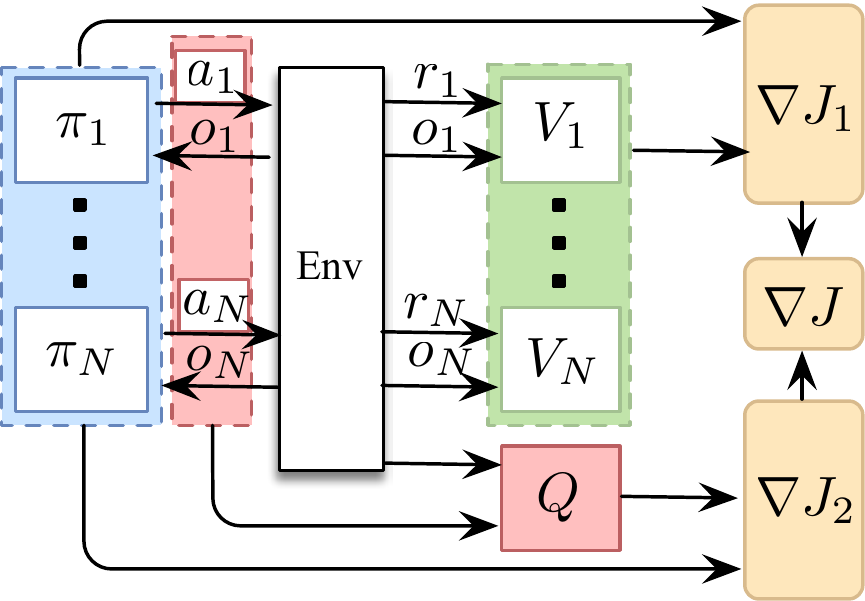}
		\caption{Stage 2: Intelligent World}
	\end{subfigure}
	\caption{Two-stage curriculum learning structure. The decentralized critic V has the same input and network structure as the policy network. The centralized critic Q has full information of all agents as input and is connected to two FC layers with 128 units.}
	\label{fig: training}
\end{figure}

\subsection{Different Driving Behaviors}

We aimed at training a single policy network that can be used on roads with different priorities and is capable of generating different diving behaviors directly according to distinct behavioral parameters. In this way, the multi-agent driving environment will contain agents with various driving styles, which will encourage each agent to select adaptive strategies when interacting with different types of drivers. 

To achieve this goal, we defined the reward function as $r(s,a,b)$, such that, besides the state and action pair, the reward also depends on the priority level $ b_{prio}$ and the driver type $b_{type}$. Specifically, the reward function for each agent is composed of two sub-reward functions that are related to the driving behavior : $r_{finish}$ and $r_{collide}$. Each agent will be assigned with a one-time reward if it safely drives through the merging scenario and the reward value depends only on the driver type $r_{finish} = f_1(b_{type})$. For example, a small final reward will encourage the agent to finish the task faster than a large reward due to the discount factor $\gamma$, which induces a less cooperative driver type. If two agents collide, a negative reward will be assigned to each agent according to the road priority, namely, $r_{collide} = f_2(b_{prio})$. For instance, if an agent driving on the main lane collides with a merging agent, the latter will take more responsibility since the main-lane car has the right-of-way. 

\subsection{Curriculum Learning}
It is difficult and inefficient to let the agent concurrently learn adaptive and interactive driving behaviors especially under multi-agent setting, where the state and action space grow exponentially with the number of agents. Therefore, 
we introduced a two-stage training process which utilized the curriculum learning \cite{curriculum} strategy. 
We successively placed the agent in a \textit{Robot World} and a \textit{Intelligent World} during the first and second training stage, with training episodes of 20k and 50k respectively. The discount factor $\gamma$ is selected as 0.9 and the weighting factor $\alpha$ is set to 0.7. 

For the first stage, we randomly placed an agent in a \textit{Robot World} environment, where multiple robot cars are driving on the road with predetermined maneuvers. These robot cars are strictly following the driving rules $\pi_{robot}$, where they never yield to the merging car and always give ways to the main-lane car. In order to drive safely in such robot world, the agent has to learn how to strictly follow the rules as well while having different driving behaviors. Under this stage, we trained an actor $\pi_1$ and decentralized critic $V^{\pi_1}$ according to the policy gradient $\nabla J_1$, as shown in Fig.~\ref{fig: training}(a).

\begin{figure}[htbp]
	\centering
	\includegraphics[scale=0.40]{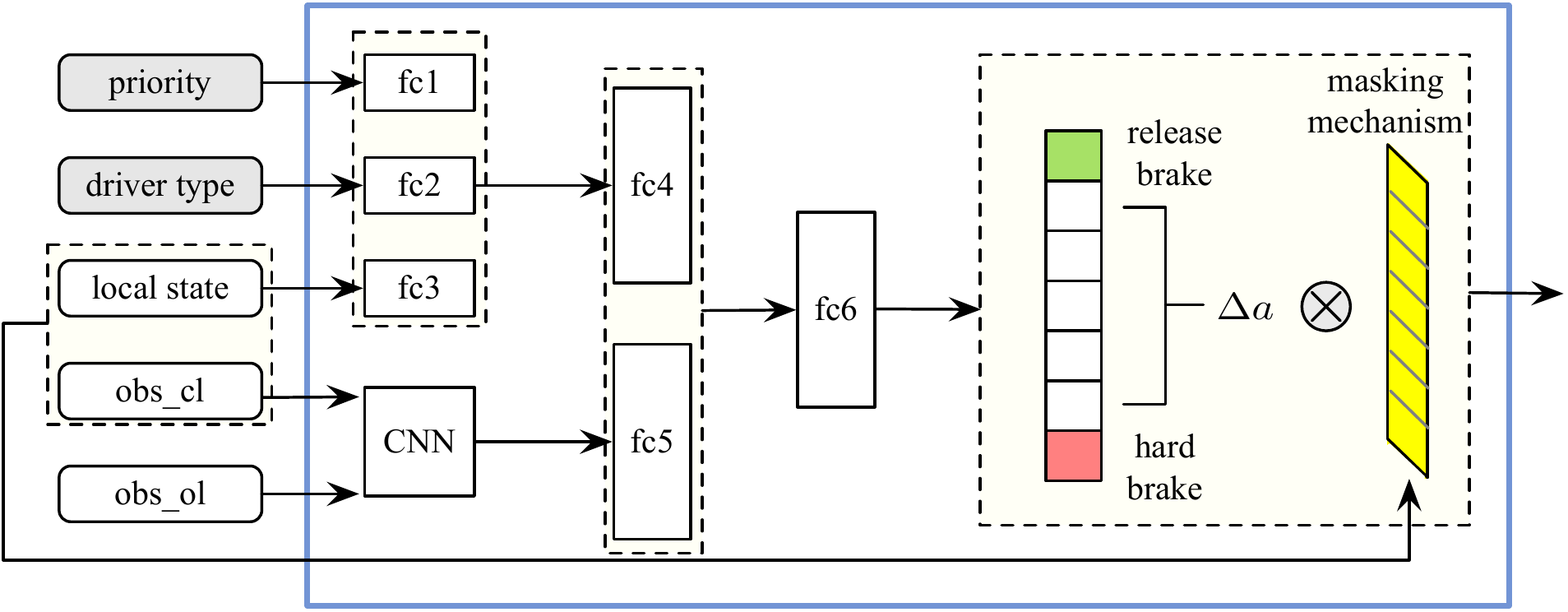}
	\caption{Policy network with masking mechanisms. We used 32 units in fc1, fc2, and fc3 layers; the observation input are processed by a convolution layer with 2 filters of size 3x30, stride 1x1; fc4, fc5, and fc6 all have 64 neurons. We used ReLU as the non-linear activation function for all layers. }
	\label{fig:policy_net}
\end{figure}
For the second stage, we randomly placed eight agents on the road with the same policy trained in Stage1, where no more robot cars are included and agents with different behaviors will learn how to interact with each other intelligently. Based on the pretrained policy and decentralized critic, we added a centralized critic $Q^{\pi}(s,\bm{a},\bm{b})$ and continued training $\pi$, $V$, $Q$ for interactive behaviors using the combined gradient $\nabla J$ and the two losses $\mathcal{L}(\phi_1)$, $\mathcal{L}(\phi_2)$. The structure is shown in Fig.~\ref{fig: training}(b).




\subsection{Masking Mechanism}
Instead of exploring actions that will definitely cause an accident or brake the rules, we expect our policy to directly learn how to make decisions on the tactical level. Therefore, a masking mechanism is applied to the policy network, where three kinds of masks are considered based on vehicle kinematics ($\mathcal{M}_k$), traffic rules($\mathcal{M}_r$), and safety ($\mathcal{M}_s$) factors. The direct effect of this is that before choosing the optimal action, we apply the masking mechanism so that only a subset of feasible actions can be selected. The final mask is equal to the union of all three masks: $\mathcal{M}=\mathcal{M}_k \cup \mathcal{M}_r \cup \mathcal{M}_s$. 

The kinematic mask keeps agent's acceleration and jerk in feasible ranges; the traffic-rule mask prevents the agent from exceeding the road speed limit; and the safety mask let the agent keep a safe distance from its front car, so that there will be enough reaction time when emergency happens.



\section{Experiments}
\subsection{Environments}
We developed a driving simulator\footnote{The simulation environment and some testing results can be found on \url{https://youtu.be/2CTTFHDW1ec}. Detailed descriptions of the simulation environment as well as a more intuitive summarization of the paper can be found in the video attachment.} specifically for merging scenarios, where the road geometry, speed limit, and road priority level can be randomly defined. During training, the total number of agents on the road, traffic densities, agents' initial states, and the time each agent enters the scene are randomly assigned, which provide enough stochasticity to this merging problem. 



\subsection{Implementation Details}
\subsubsection{Basic Information}
We constructed two roads with 250 meters each, where each of them can be defined as either the main lane or the merge lane with the speed limit of 20m/s or 15m/s. Similar to the real driving scenario, we allow the merging vehicle to accelerate and exceed the speed limit when its distance to the merging point is below 100m. These road constraints are applied through the rule-based mask $\mathcal{M}_r$.

Moreover, we considered some commonly used kinematic constraints for our agents by adopting the $\mathcal{M}_k$ mask, where the acceleration is within $[-2,2]m/s^2$ and the jerk range is $[-1,1]m/s^3$. The Intelligent Driver Model (IDM) is also utilized in order to let the agent keep a safe distance from its front vehicle, which is implemented in the safety-based mask $\mathcal{M}_s$. By using IDM, the maximum acceleration can be calculated as:

\begin{equation}
\begin{split}
&\dot{v_\alpha} = a \Bigg( 1 - {\Big( \frac{v_\alpha}{v_0} \Big)}^{\delta} - {\Big( \frac{s^\ast(v_\alpha, \Delta v_\alpha)}{s_\alpha} \Big)} ^2 \Bigg)\\
&s^\ast(v_\alpha, \Delta v_\alpha) = s_0 + v_\alpha T + \frac{v_\alpha \Delta v_\alpha}{2\sqrt{ab}} ,
\end{split}
\end{equation}
where $v_0$ denotes the velocity limit, $s_0$ denotes the minimum distance between two vehicles, which is set to 2m in this problem, $T$ denotes the time headway, and $a, b$ are the maximum and minimum acceleration respectively. 

\begin{table*}[ht]
	\begin{minipage}[c]{0.4\linewidth}
		\centering
		\begin{tabular}{p{1.5cm} p{1.1cm} p{1.1cm} p{1.1cm}}
			\toprule 
			& Success (\%) & Perfect Success (\%) & Avg. Speed (m/s) \\
			\midrule
			IDAS & \textbf{100} & \textbf{95} & \textbf{17.51} \\ 
			\midrule
			IDAS-V & 93 & 64 & 13.76 \\ 
			\midrule
			IDAS-direct & 77 & 23 & 15.44 \\ 
			\midrule
			\midrule
			IDM & 80 & 12 & 17.01 \\ 
			\midrule
			FSM+IDM & 94 & 57 & 14.45 \\
			\bottomrule
		\end{tabular}
		
		%

	\end{minipage}\hfill
	\begin{minipage}[c]{0.62\linewidth}
		\centering
		\includegraphics[scale=0.63]{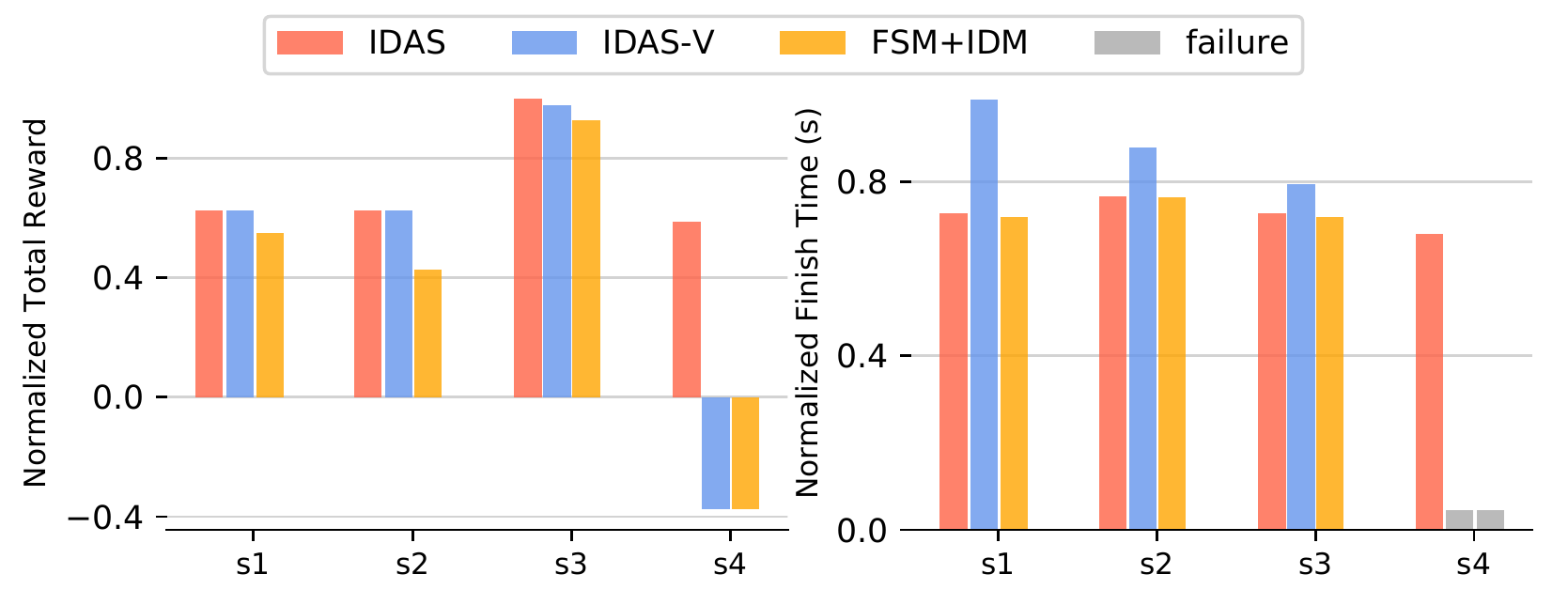}
	\end{minipage}
	
\captionof{figure}{The left table shows the overall comparison over all methods. The \textit{success rate} describes the number of success cases in all testing scenarios; the \textit{perfect success rate} denotes the number of times that all agent succeed without any reward dedcution under all testing scearios; the \textit{avg speed} calculates the average speed for all agents in the \textit{perfect success} cases. The right two histograms compare the normalized reward and average finish time among selected methods over four scenarios where detail descriptions can be found in Section VI.B. When two vehicles crash into each other, we consider it as the failure case and color in gray.}
\end{table*}

\subsubsection{Policy Network}
The policy network structure is shown in Fig.~\ref{fig:policy_net}, which has five types of input and seven actions. Each agent has a priority number, either 0 or 1, according to its current state, where 1 have higher priority than 0. Two agents will have the same road priority when their numbers are both 0 or 1. Note that an agent's priority will change during the entire process if its road priority changes. (e.g. merge into the main road from an on-ramp. )

The driver type is a continuous number ranging from -2 (less cooperative) to 2 (really cooperative), which will be randomly selected and fixed for each agent. Every agent is able to receive its own local state information including the current speed, acceleration, and distance to the merging point. According to the sensor range for actual autonomous vehicles, we assume the visibility of each agent is 100m in front and back, with 0.5$m/cell$ resolution, for both its current lane $(obs\_cl)$ and the other lane $(obs\_ol)$. Therefore, the observation input for each lane is a discretized one-dimensional observation grid centered on the agent, with two channels: binary indicator of vehicle occupancy and the relative speed between other vehicles and the agent. 

All agents have the same action space, consisting of five discretized values for change of acceleration $\Delta a $, one hard-brake action, and one release-brake action. Although the action space is discrete, by assigning small values to $\Delta a$, the corresponding acceleration can become roughly continuous. Also, in self driving cars, a motion control module will convert the discretized results of the behavior planner into continuous acceleration by applying algorithms like Model Predictive Control (MPC) at a higher frequency (100Hz). Moreover, by using $\Delta a $ as our action instead of $a$, we can have a wide range of actual control input sent to the vehicle while having a low-dimensional action space, which can enhance the training efficiency.

Based on the jerk range and the 5Hz decision-making frequency, we have $\Delta a = \{-0.2,-0.1,0,0.1,0.2\} m/s^2$. Although the acceleration is usually between $-2$ to $2$ $m/s^2$, to insure safety, the vehicle should be able to brake hard during emergency situations. Therefore, we added a hard-brake action that enables the agent to deceleration at $4 m/s^2$ when necessary. A release-brake action, with the acceleration of $0 m/s^2$, will be available directly after a hard-brake action in order to let the agent recover from a large negative acceleration and speed up quickly. Notice that we no longer need to include the jerk constraint in the $\mathcal{M}_k$ mask since it is already satisfies in the action space.


\subsubsection{Rewards}
The designed rewards can be divided into two categories.
The first two rewards, $r_{finish}$ and $r_{collide}$, correspond to the \textit{microscopic-level} that influence the dynamics of each agent; the remaining two \textit{macroscopic-level} rewards take all road participants as a whole and focus on the traffic flow characteristics. Specifically, $r_{impede}$ penalizes agents that cause other agents to hard brake; $r_{flow}$ penalizes every sudden brake happens inside the merging zone which is defined as the circular area with 50m radius around the merge point. Since every agent should be responsible for a bad traffic flow, the last reward will be considered only in the joint reward instead of as a penalization for any agent. 
\begin{itemize}
	\item $r_{finish}: f_1 = (15*b_{type} + 50)/4$, $b_{type}\in[-2, 2]$
	\item $r_{collide}: f_2 = -5*b_{prio} - 5$, $b_{prio} \in \{0,1\}$
	\item $r_{impede}: -5$
	\item $r_{flow}: -1$
\end{itemize}


\section{Evaluation and Results}
In this section, we first compare the overall performance of our method with other methods. Then, we select four representative scenarios to assess each approach through further inspections. Finally, detailed evaluations are provided to illustrate the capabilities of the proposed approach.

\begin{figure*}
	\centering
	\includegraphics[width=0.90\textwidth]{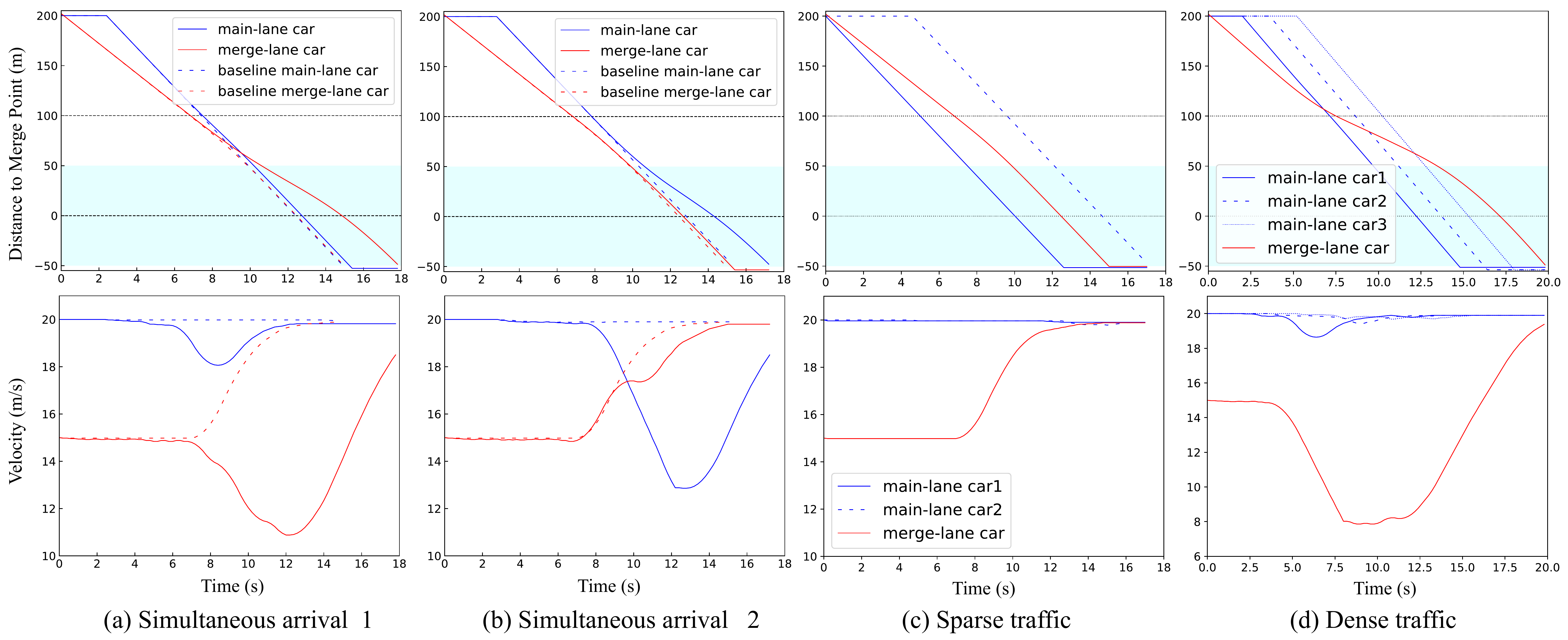}
	\caption{Selected representative scenarios illustrated in the position-time domain and via the velocity profile. The blue areas are the merging zone as defined in Section IV.B(c). For a uniform comparison, the driver type for the main-lane car and merge-lane car is -1 and 1 respectively. All agents are operating on our learned policy.}
	\label{fig:interaction_plot}
\end{figure*}


\subsection{Overall Performance}

We selected several criteria for validation and compared our method with the following four approaches. The first two are used for ablation study and the last two are commonly-used methods for decision-making problems. 

1) \textit{IDAS-V} : It only uses the policy generated from the first training stage. 


2) \textit{IDAS-direct}: We directly trained the policy (with 70k training episodes) using our algorithm structure without utilizing curriculum learning.

3) \textit{Intelligent Driver Model (IDM)} : The agent will follow its front car using the IDM model and when no cars in front, it will drive at the road speed limit.

4) \textit{Finite State Machine (FSM) + IDM}: By incorporating FSM, we manually set how each agent should interact with other agents on the other lane. Specifically, the agent will first examine if there is any vehicle on the other lane that has similar distances to the merge point as itself. If there is such vehicle, the agent will select that car as another front vehicle besides its actual front vehicle and then utilize IDM to follow both of them. 


%

We randomly generated 100 scenarios that cover various road situations for testing and the general performances of the above methods are shown in the left table in Fig. 4. Note that when used in the real world, the learned policy will only be applied by the ego car. Therefore, to compare these methods, we first randomly selected a vehicle in the scene to be our agent and then apply each of the selected method to the chosen agent. At the same time, instead of applying a simple driving model to unselected agents, we used the learned IDAS policy on them with different assigned driver types, which provides sufficient randomness to the testing environments. Moreover, to make a fair comparison, all these methods are equipped with the masking mechanism. 

As can be seen in the table, our approach outperforms other methods in all evaluated metrics and achieves zero collision.
Notice that the perfect success rate for IDAS-V is low even for a high success rate. This is because by strictly following the road priorities, the merging agents always yield to the main-lane traffic to avoid potential collision while fail to consider the overall traffic flow and thus impede other agents behind. The success rate and average speed value for the IDAS-direct are lower than IDAS that uses the curriculum learning strategy, which is due to the fact that the network tries to learn everything at once and fails to acquire the desired behaviors within insufficient learning episodes. The two traditional methods, IDM and FSM+IDM, have reasonable success rates but their perfect success rates are relatively low. The reason is because these methods strictly follow the hand-designed rules and cannot have adaptive strategies when dealing with varying situations. Therefore the ego agent will count on other road entities to react to its maneuvers without considering the discomforting of other participants and any negative influences to the traffic. 
Such behavior will lead to serious collision if other drivers do not pay much attention to the ego car especially when they have the right-of-way.

\subsection{Representative Scenes}
Based on the evaluation results in the previous section, we selected the best three methods to further examine their performances. Four scenarios that involve high interactions are selected, where two agents are placed in the scene with one agent on each road and the details are shown below:

1) s1: $b_{prio}^1 \neq b_{prio}^2$, $b_{type}^1 \neq b_{type}^2$, $t_{mpt}^1 > t_{mpt}^2$,

2) s2: $b_{prio}^1 \neq b_{prio}^2$, $b_{type}^1 \neq b_{type}^2$, $t_{mpt}^1 < t_{mpt}^2$,

3) s3: $b_{prio}^1 \neq b_{prio}^2$, $b_{type}^1 = b_{type}^2$, $t_{mpt}^1 = t_{mpt}^2$,

4) s4: $b_{prio}^1 = b_{prio}^2$, $b_{type}^1 \neq b_{type}^2$, $t_{mpt}^1 = t_{mpt}^2$, \\
where $t_{mpt}^n$ denotes the time that car $n$ will arrive at the merge point if no interaction involved. Note that even if two cars do not arrive at the merge point simultaneously as in $s1$ and $s2$, their time difference of arrival is bounded by a small value $\epsilon$, $|t_{mpt}^1 - t_{mpt}^2| < \epsilon$, to ensure possible interaction. Different from the previous evaluation, we applied the same decision making approach to both agents. However, since the FSM+IDM method cannot have different driver types and will have the same driving behavior across all situations, we apply the IDAS policy to one of the agents in order to test how this decision maker react in varying environment.  


According to the histograms in Fig. 4, IDAS can achieve the highest total reward while maintaining high merging efficiency in all testing scenarios. The other two methods,on the other hand, either have long finishing time or low reward values. Also, both methods fail in the last scene where two roads have same priority. This is because when we set both roads as the main-lane that have equally high priority, IDAS-V lets two agents strictly follow the driving rules and without considering other road's situation, which leads to collision. Note that FSM+IDM can only handle interaction when the agent has a front car on the other road and will not decelerate when no cars in front. Therefore, if any vehicle drives slightly behind the ego car or by its side, it will leave insufficient reaction time for the other agent and thus collision happens.



\subsection{Detailed Evaluations of IDAS}
\subsubsection{Adaptive Strategies}

To demonstrate that the agent indeed learns to adapt its strategy during different scenarios, we visualized the testing results in the position-time domain as well as through the velocity profile, which are shown in Fig.~\ref{fig:interaction_plot}. In all four selected scenarios, both agents will apply the IDAS policy and one road is defined as the main-lane while the other road as the merge-lane that has lower priority. 


According to the result, although each agent applies the same policy across these scenarios, it exhibit different driving strategies when dealing with different road situations. The merging car yields to the main-lane car when necessary in (d), and accelerates to merge when enough gap is available on the main road in (c). An interesting phenomenon we found in (d) is that although the main-lane agent has the right-of-way, each of the three vehicle tends to slightly press the brake when it has a similar distance to the merge-point as the merging agent. Such behavior resembles actual human drivers, where we usually, for safety purpose, decrease our speed when passing through a road with traffic.

When both agents reach the merging zone simultaneously, Fig.~\ref{fig:interaction_plot} (a) and (b), they begin to negotiate to avoid potential conflict. The difference between these two scenarios is that the longitudinal distance between the two cars is larger in (b) than in (a). In both situations, two agents first start to slightly decelerate before reaching the merging zone, which is similar to what human driver will do in order to prepare for emergencies. Then, in case (a), the merging car continues to decelerate due to its higher cooperativeness level and lower priority, which enables the  main-lane car to accelerate since it detects the yielding intention from the merging car. 

Contrarily, in case (b), although it has low priority, the merging car notices that it is ahead of the main-lane car with an acceptable distance and may have a chance to merge first. Thus the car on the merge lane begins to accelerate, which causes the main-lane car to decelerate. Instead of keep accelerating, the merging car remains its speed for a while to infer the behavior of the main-lane car and then continuous to accelerate as it detects that the other car has the willingness to yield. In contrast, if two vehicles ignore each other and fail to interact in both scenarios, a collision will happen as shown in the dashed lines. More testing results can be seen in the provided video link.

\subsubsection{Different Driver Types}
To examine whether the learned policy can generate different driving behaviors correspond to different driver types, we fixed all parameters in the scenario considered in Fig.~\ref{fig:interaction_plot}(a) except for the driver type of the merging car. As shown in Fig. 6 (a), by continuously increasing $b_{type}^{merge}$ from -2 to 2, the average merge time increases. Also, in Fig.6 (b), for different driver types, the merging car's velocity profile has substantial variations, which slightly influence the velocity profile of the main-lane car as well. We noticed that as the merging agent becomes more cooperative, it decelerates for a longer time to ensure a safer merging. Moreover, after a certain threshold in Fig. 6(b), the merging car discontinue to decelerate since it learns to merge efficiently and not impede the traffic flow.

\section{Conclusions}

In this paper, we proposed an approach called IDAS to solve decision-making problems under merging scenarios for autonomous vehicles. By introducing the concept of driver type and road priority, we expect the self-driving cars to automatically infer the behaviors of other drivers during interaction and leverage their cooperativeness to strategically handle different merging situations. We solved this problem under the multi-agent reinforcement learning setting and proposed a double critic containing a decentralized value function as well as a centralized action-value function. The curriculum learning and masking mechanism is also applied to the policy network to increase the learning efficiency. According to the evaluation results, our approach outperforms other selected methods in terms of the success rate and merging efficiency.
Moreover, by further assessing IDAS, we concluded that the learned policy can successfully generate distinct driving behaviors and make adaptive strategies during interaction. Although the learned policy can achieve zero collisions in all tested scenarios, one drawback of RL-based methods is that agent's safety cannot be fully guaranteed especially in the real environment with high inherited uncertainty. Therefore, for future work, we will focus on how to transfer the learned policy from the simulation to the real-world with sufficient safety considerations. Based on the current algorithm structure, we will also extend the problem to more complicated scenarios with unknown merge point.

\begin{figure}[htbp]
	\centering
	\includegraphics[scale=0.28]{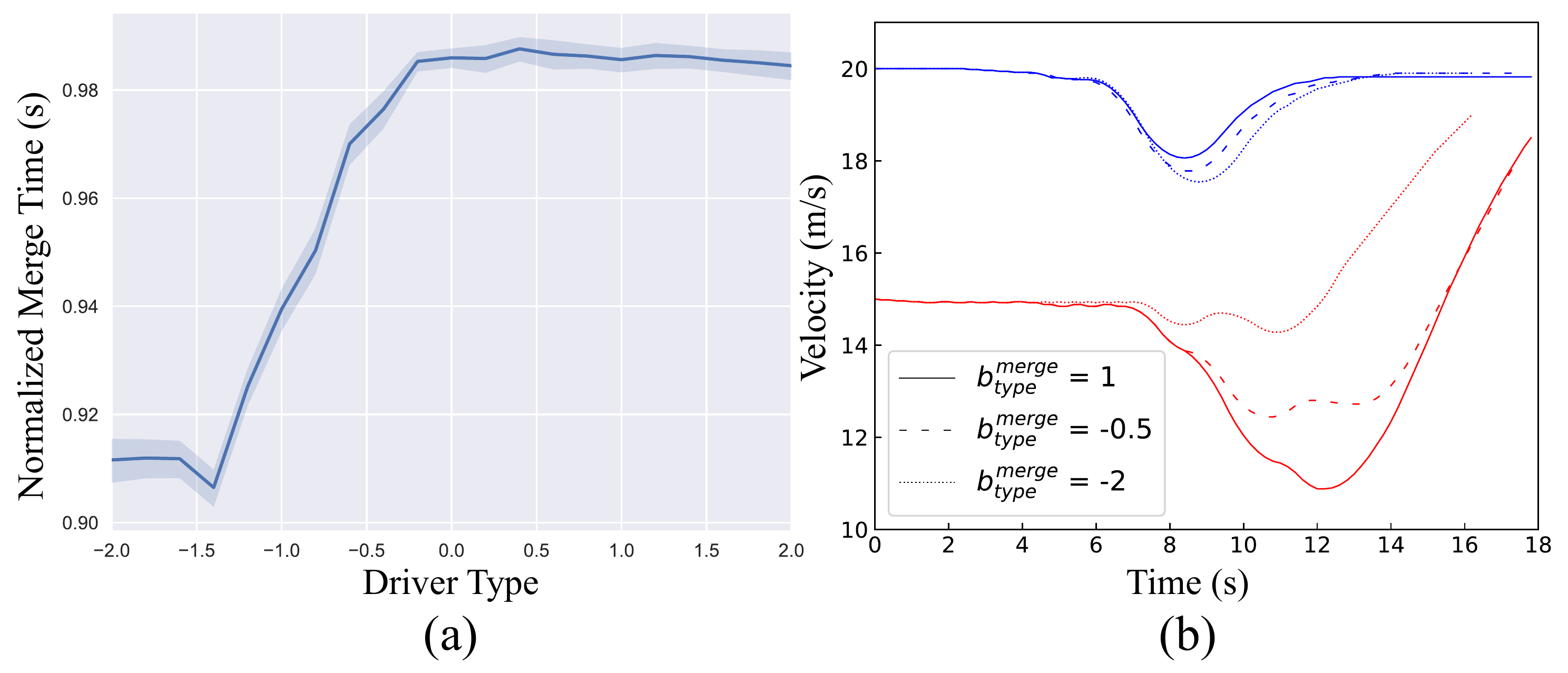}
	\caption{Evaluation of different driver types.}
	\label{fig:adaptivity}
\end{figure}




\addtolength{\textheight}{-10cm}   








\balance

\bibliographystyle{IEEEtran}
\bibliography{IROS2019_ref}

\end{document}